# Explanation as Question Answering based on Design Knowledge

**Ashok Goel, Vrinda Nandan, Eric Gregori, Sungeun An, and Spencer Rugaber**
Design Intelligence Laboratory, School of Interactive Computing, Georgia Institute of Technology
ashok.goel@cc.gatech.edu, vrinda@gatech.edu, egregori3@gatec.edu, sungeun.an@gatech.edu, spencer@cc.gatech.edu

**Abstract**

Explanation of an AI agent requires knowledge of its design and operation. An open question is how to identify, access and use this design knowledge for generating explanations. Many AI agents used in practice, such as intelligent tutoring systems fielded in educational contexts, typically come with a User Guide that explains what the agent does, how it works and how to use the agent. However, few humans actually read the User Guide in detail. Instead, most users seek answers to their questions on demand. In this paper, we describe a question-answering agent (AskJill) that uses the User Guide for an AI-based interactive learning environment (VERA) to automatically answers user's questions and thereby explains VERA's domain, functioning, and operation. We present a preliminary assessment of AskJill in VERA.

## Introduction, Background and Goals

AI research on explanation has a long history that dates at least as far back as the rise of expert systems in the 1960s, e.g., DENDRAL (Lindsay et al. 1993). Mueller et al. (2019) provide a recent and comprehensive review of this research. One of the key ideas to emerge out of this early research was the importance of the explicit representation of knowledge of the design of an AI system (Chandrasekaran & Swartout 1991; Chandrasekaran & Tanner 1989): an explicit representation of the design knowledge of an AI system enables generation of explanations of the tasks it accomplishes, the domain knowledge it uses, as well as the methods that use the knowledge to achieve the tasks. This raised the question of how this design knowledge can be identified, acquired, represented, stored, accessed, and used for generating explanations. One possible answer was to endow the AI agent with meta-knowledge of its own design (e.g., Goel et al. 1996). However, much of AI research on expert systems collapsed by the mid-1990s.

Starting in the 1970s, AI research on explanation also encompassed intelligent tutoring systems (Buchanan 2006). Indeed, in the 1990s, given the collapse of AI research on expert systems, the focus of AI research on explanation shifted to intelligent tutoring systems. Unlike the design stance towards explanations adopted by the research on expert systems, research on tutoring systems took a strongly human-centered perspective towards the generation of explanations. This view emphasized the users and the uses (or contexts) of explanations (e.g., Woolf 2007). For example, Graesser, Baggett & William's (1996) describe question-answering as a basic mechanism of generation of explanations in intelligent tutoring systems, where the answers to the questions meet the requirements and expectations of the human users; Aleven & Koedinger (2002) present explanations of reasoning as a source of new knowledge and learning for the users. However, much of this work perhaps lay a little outside mainstream AI research.

Over the last five or six years, explanation has again entered mainstream AI research (e.g., Gunning & Aha 2019). This is in part because of advances in machine learning, such as deep learning, that have refocused attention on the need for interpretability and explainability of internal representations and processing in AI agents in general. In this paper, we take the two ideas from explanations in expert systems and tutoring systems mentioned above as our starting points: (1) use of the knowledge of the design of an AI agent as the basis for generating explanations and (2) human-centered question-answering as the basic mechanism for generation of explanations. We add a third idea to this mix: given that most practical AI agents, for example almost all intelligent tutoring systems, come with a User Guide that contains knowledge about the domain, design and operation of the agent (Ko et al. 2011), might the User's Guide act as a basis for generating explanations? Note that by definition, the User Guide contains information about many types of explanations that users want. For example, a User Guide for an AI agent typically contains information about the domain of the agent, the vocabulary for representing the domain knowledge, the functions of the agent (what it does), the structure of the agent (its components), the behaviors of the agent (the internal processing of the agent), as well as the operation of the agent (how to use the agent). However, few humans actually read the User's Guide in any detail (Rettig 1991; Novick and Ward 2006; Mehlenbacher et. al. 2002). Instead, most users want answers to their questions on demand, as and when needed. Thus, (3) we propose to use the User Guide to generate answers to users' questions. To put it another way, we recast explanations in AI agents in intelligent learning environments as an interactive User Guide for answering users' questions. A corollary here is that we seek to identify the design knowledge a Users' Guide must contain to act as a basis for generating explanations.

In this paper, we describe the use of a question-answering agent (called AskJill) for generating explanations about an interactive learning environment (named VERA) based on the latter's User's Guide. AskJill is intended to automatically answer users' questions and thereby explain VERA's domain, functioning, and operation. We present a preliminary formative assessment of AskJill in VERA.

## VERA, An Interactive Learning Environment

The VERA project addresses the issues of accessibility, achievability, and quality of online education. Residential students in higher education have access to physical laboratories, where they conduct experiments and participate in research, thus discovering new knowledge grounded in empirical evidence and connecting it with prior knowledge. Online learners do not have access to physical laboratories, which impairs the quality of their learning. Thus, we have developed a Virtual Experimentation Research Assistant (VERA for short) for inquiry-based learning of scientific knowledge (An et al. 2020, 2021): VERA helps learners build conceptual models of complex phenomena, evaluate them through simulation, and revise the models as needed. VERA's capability of evaluating a model by simulation provides formative assessment on the model; its support for the whole cycle of model construction, evaluation, and revision fosters self-regulated learning. Given that even residential students have only limited access to physical laboratories, VERA is also useful for blended learning.

VERA is available online (http://vera.cc.gatech.edu) for free and public use. Although VERA is useful for any agent-based domain, for the specific domain of ecology, we have integrated it with Smithsonian Institution's Encyclopedia of Life (EOL) that is available as an open-source library and software (EOL; www.eol.org; Parr et al 2016). EOL's TraitBank supports ecological modeling in VERA in several ways: it provides (i) the ontology of conceptual relations for conceptual modeling, (ii) knowledge of specific relations among biological species in a given ecological system, and (iii) the parameters for setting up the simulations. Given that the space of simulation parameters can be very large, and a learner may not know the "right" values for the parameters, once the learner sets up the conceptual model using the EOL digital library, VERA further uses EOL's knowledge of biological species to directly set initial values of the simulation parameters. The learner may then tweak the parameter values and experiment with them..

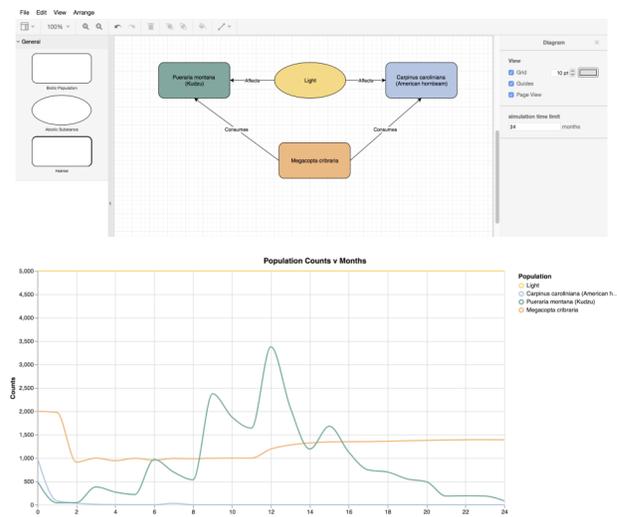

Figure 1. (a) An example of a conceptual model (the top half of the figure) and (b) its agent-based simulation automatically generated by VERA (the bottom half).

Figure 1 illustrates the use of VERA to model the impact of a kudzu "bug" to moderate the impact of kudzu, an Asian invasive species, on the American hornbeam, a kind of tree common in the eastern half of the United States. In Figure 1(a), the learner interactively builds a conceptual model, and in Figure 1(b) VERA illustrates the results of an agent-based simulation of the model. In this case, the simulation results show that because of the introduction of the kudzu bug, the population of kudzu will decline over time and the American hornbeam will survive.

At present, VERA provides four of the most common types of ecological phenomena: predator-prey, exponential growth, logistic growth, and competitive exclusion. These four types of models together account for a vast majority of ecological phenomena. VERA uses agent-based simulations to provide formative assessment on the conceptual models. Note that VERA automatically spawns agent-based simulations from the conceptual models: an AI agent inside VERA understands enough of the syntax and semantics of both the conceptual models and the agent-based simulations that it can automatically spawn the latter from the former. This is another example of learning assistance in VERA. This learning assistance enables student scientists as well as citizen scientists to model complex phenomena without requiring expertise in the mathematics or the mechanics of agent-based simulations. Further, VERA's support for the whole cycle of model construction, evaluation, and revision fosters self-regulated learning.

In 2019, Smithsonian Institution started providing access to VERA directly through the main page on its EOL website (www.eol.org). Now EOL's many users can try out ecological models of several species available in EOL. These species are modeled in VERA using the data directly retrieved from EOL such as lifespan, body mass, offspring count, reproductive maturity, etc. This means that the hundreds of thousands of EOL users across the world, including learners and teachers, as well as citizen and professional scientists now have direct access to VERA. This opens up the potential for online learning in open science. It also makes explanations of VERA's domain, functioning and operation critically important.

## User Guide in VERA

VERA's User Guide and its table of contents is available on VERA's website under the Help section. It includes a written guide describing how users can build and simulate ecological experiments on VERA, the tool's expected behavior, explanations for the vocabulary terms and parameters users can manipulate, and screenshots showing the tool's structure (screens and buttons). Specifically, the 27-page User Guide covers an introduction to VERA, system requirements, steps to access the tool, general approach to build and evaluate a conceptual model of an ecological system, how to use the VERA tool for modeling and simulation (including steps to create a project describing a phenomena and associated models to test various hypotheses), how to use the model editor to manage constituent components and their relationships, how to simulate a model, how to edit model parameters to manipulate results, and ways to get help on the tool.

The User Guide provides illustrative descriptions of the user's workflow on VERA. For example, in its "Getting to know the model editor" section, the User Guide provides an example of a "starter" conceptual model of a simple ecosystem composed of wolves, sheep, and grass, to walk the user through the steps needed to create a the "biotic population" components for each of the three populations. It also shows the user how to define the ecological relationships (destroys, produces, consumes, becomes, affects, can migrate to) between each set of components (e.g. wolves "consume" sheep, sheep "consume" grass), and simulate the model. The User Guide describes how users can set up, start, stop, reset the simulation and export resulting graphs. The User Guide also provides example parameter values showing how parameters can be initialized (Smithsonian's EOL supplies default values) and tuned (provides tuning values) to get the desired population behavior (shows resulting graphs for reference) in the simulation. Last but not the least, the User Guide provides definitions and explanations for commonly used model components (e.g. biotic substance, abiotic substance, and habitat) and their associated simulation parameters (e.g. some parameters for a biotic substance are lifespan, carbon biomass, minimum population, etc.).

## AskJill, A Question-Answering Agent

AskJill is a question-answering agent embedded in the VERA interactive learning environment that automatically answer users' questions and thereby explains VERA's domain, functioning, and operation. When a user first logs-in on the VERA website, AskJill welcomes them and prompts them to ask their questions about VERA.

The user can type their questions into the AskJill question-answering interface (integrated into the VERA website). AskJill provides precise and correct answers to questions within the User Guide domain within a few seconds. Figure 2 shows a couple of examples of question-answering in AskJill.

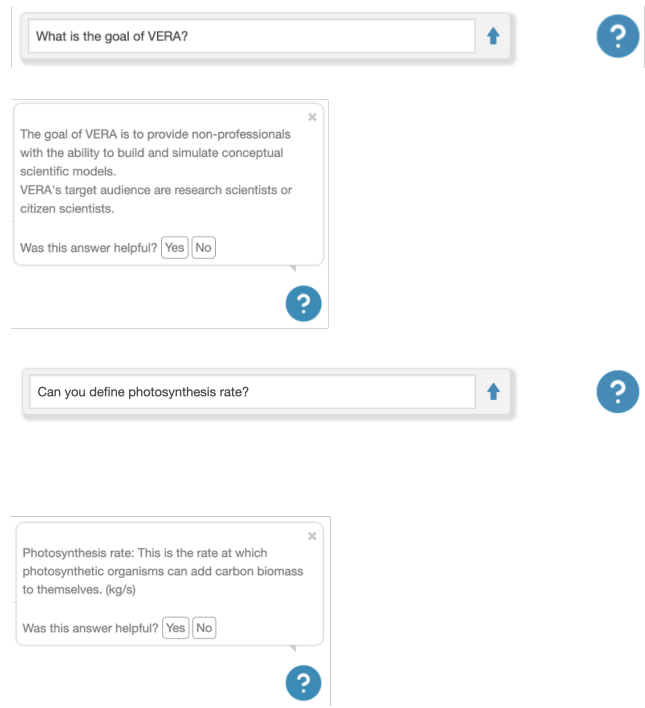

Figure 2: A couple of user questions to AskJill for VERA and AskJill's answers to the questions.

### Data Flow of an Answer to a Question

Figure 3 shows AskJill's question-answering data flow diagram. After a user asks a question in VERA's AskJill interface, it is sent to the AskJill system via a REST API. Inside AskJill, the question is parsed, and then sent to a 2D hybrid classification system. The system contains an innovative 2-stage classification process (Goel, 2020). The first is a pre-trained NLP-based intent classification layer that classifies each new question into one of the existing question categories based on user intents. The second is the semantic processing stage that converts the intent to a rule-based query. From the 2D hybrid classification system, a query is sent to the VERA's design knowledge (database) and a response is generated. The response generation system retrieves the associated query response and returns an answer if its confidence value exceeds the minimum threshold (97%). Finally, the dialogue management system post-processes the resulting response, converts it into a "human-like" natural language answer, and sends it back to AskJill in the VERA user interface. After answering, AskJill prompts the user to provide feedback, asking "Was this answer helpful", and stores the user feedback in her database. That feedback is subsequently used for retraining the agent in the future. If AskJill is unable to answer a question, it gently redirects the conversation into AskJill's domain of competence by suggesting alternate Q/A topics associated with questions the agent is trained on.

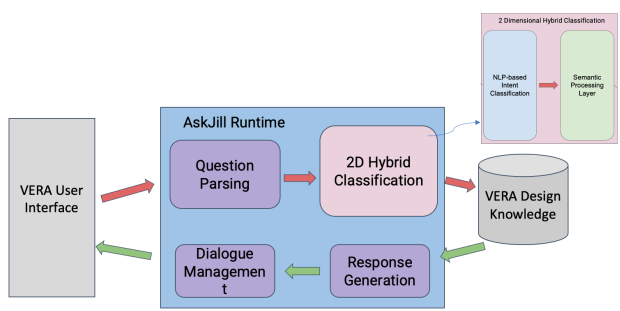

Figure 3: AskJill in VERA: Question-answering data flow diagram

**Agent Smith: Building AskJill for VERA's User Guide**

AskJill evolved from our earlier work on the Jill Watson project (Goel & Polepeddi 2018) that automatically answered students' questions on discussion forums of online and hybrid classes. Agent Smith is an interactive generator for generating Jill Watson teaching assistants for different classes (Goel 2020): it combines knowledge-based AI, supervised machine learning, and human-in-the loop machine teaching for training a Jill Watson assistant for a new class. Since AskJill for VERA's User Guide has the same architecture and algorithms as the original Jill Watson for class syllabi, we were able to reuse the Agent Smith generator to build the AskJill for VERA. Reusing the Agent Smith technology allows us to train, retrain and generate AskJill agents based on VERA's User Guide efficiently and easily. Agent Smith also enables the extension of AskJill to other domains beyond ecology. AskJill's VERA domain is defined by an ontology for the VERA User Guide and encoded in the form of unique question templates related to goals, getting started, definitions, and how-to pointers, simulation parameter default values, and units. Similar to previous Jill Watson applications, Smith builds a semantic memory for VERA's vocabulary, system requirements, structure, and tool behavior. It also generates a knowledge base consisting of user intents, keywords and associated answers. Agent Smith then uses supervised learning to train a classifier to generate an AskJill for VERA.

Figure 4 shows an example of the question templates used for training AskJill. Agent Smith projects the templates onto the VERA ontology and generates the training dataset. The AskJill agent uses the trained model for run-time question answering. Over time, as we collect user feedback and analyze missed questions, we can expand the training dataset and retrain AskJill enabling it to answer more and more questions. As a by-product of development, testing and training the Q&A AskJill agent for VERA, we identified definitions and parameters that were initially missing in the User Guide. We have since updated the User Guide to include those missed aspects.

Figure 4: Some examples of Agent Smith Question Templates for VERA Q&A AskJill Agent

## Evaluation of AskJill in VERA

We collected AskJill user data both during its use in an Introduction to Biology class at Georgia Tech, as well as from citizen scientists discovering VERA through Smithsonian's website or while browsing the Internet (An et al. 2020, 2021). Currently, AskJill can answer questions belonging to seven categories as shown in Figure 5. Figure 6 shows examples of a human-generated question from each question category above and as well as the answers generated by AskJill.

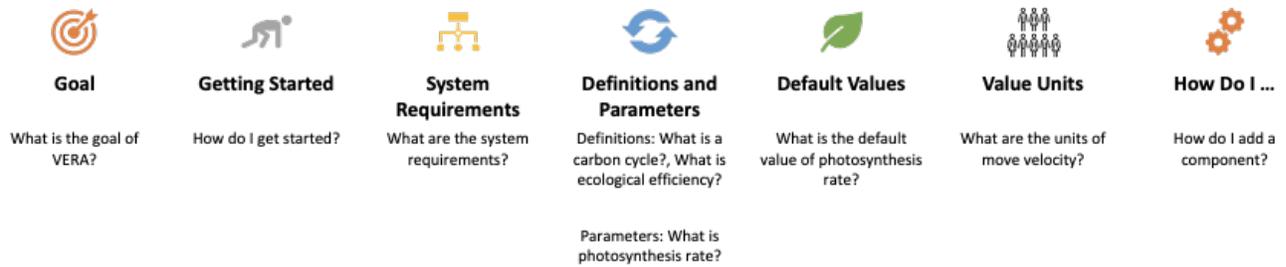

Figure 5: User Intent (question) categories on AskJill

Figure 6: Human generated questions and AskJill's agent generated answers.

The AskJill training data set has evolved over time. Originally, AskJill could resolve fewer user intents from a lower number of questions. Over time, AskJill's abilities have improved to include a larger number of intents (a total of 7 intents) and the training dataset engulfs a much larger set of user questions (3053 questions are part of the current training dataset). The expanded training dataset includes both the actual user questions, and anticipated questions from the User Guide. Originally, AskJill wasn't able to respond to a number of training dataset questions correctly. In such cases, AskJill's answers involved either redirecting the user to the User Guide or an IDK (I do not know) response encouraging the user to ask questions from to AskJill's domain knowledge. The improved version of AskJill provides semantically correct and meaningful responses for all 3053 training questions.

Given that Agent Smith automatically generated the training dataset using a combination of template questions and relevant keywords, we also tested for the grammatical correctness of the generated training dataset. Out of 3053 questions, 2907 or 95.2% were syntactically (and semantically) correct. The remaining 4.8% of questions were not grammatically correct but AskJill was still able to resolve the associated intents and answer them correctly. Figure 7 shows 100% semantic correctness of the agent generated responses to the current training question set (3053 questions). It also shows the split between syntactically correct and incorrect agent generated questions.

We have also collected a small dataset consisting of in-situ observations. Figure 7 shows a comparison of data collected from 9 users including external users as well members of our research laboratory. AskJill correctly answered 19 out of 31 unique questions for all users. We measured user satisfaction using the integrated feedback prompt (Was this answer helpful?) built into the agent's interface and validated that the users confirmed (in some cases there was no feedback) that the correctly answered responses were indeed helpful to the user. Out of the 12 questions that were not answered correctly, a majority are related to simulation parameters, simulation properties, and how-to information (only 1 out of 12 questions is related to a missed definition). We have added the related missing information to the User Guide, making it more helpful to VERA users. The closed loop process i.e. adding the information related to missed questions to the VERA knowledge domain, updating the User Guide and retraining AskJill to expand its question answering abilities has resulted in improvements to the entire VERA and AskJill pipeline.

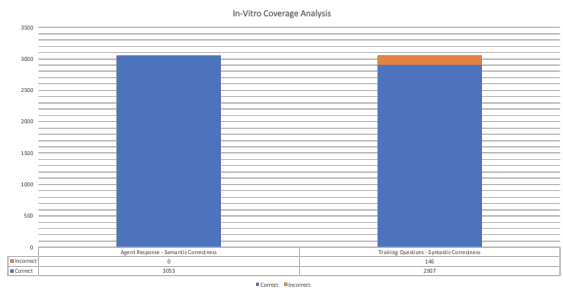

Figure 7: Agent Response Semantic Correctness and Training Question Syntactic Correctness

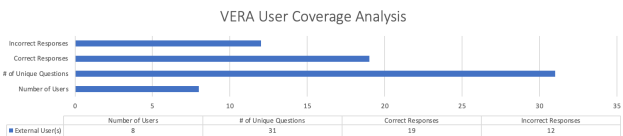

Figure 8: The bar plots show the (a) correct vs incorrect responses (includes "I do not know") responses (b) Number of unique user questions (c) Total number of users.

## Discussion

As Mueller et al. (2019) note that the right strategy for generation of explanations of AI agents depends on the context including the task and the user. AskJill generates explanations for human learners in the context of the interactive learning environment called VERA. The explanatory goal here is to help users learn about VERA's domain, design, and operation.

However, while this design approach enables general-purpose explanations, it does not afford explanations of specific instances of reasoning and action by the AI agent. Thus, this approach likely has to be complemented with an episodic approach that relies on specific cases of decision making (Langley et al. 2017). Indeed the case-based reasoning research community has developed several schemes for case-based explanation of decision making such as (Leake & McSherry 2005; Schank, Kass & Riesbeck 1994). In our own earlier work along these lines, we used meta-cases to capture derivational traces in an earlier interactive learning environment and used the meta-cases to explain the agent's decision making (Goel & Murdock 1996). A future version of AskJill may keep a derivational trace of VERA's decision making and augment its explanatory capability based on a replay of the derivational trace.

Nevertheless, even in its current form, our design-based approach provides insight into specific episodes of decision making both by explaining the vocabulary and the general mechanism of decision making. Consider, for example, decisions about the values of the simulation parameters in VERA'a agent-based simulations. AskJill can explain each simulation parameter, the role it plays in the simulation, as well as the general mechanism of the agent-based simulation.

AskJill builds our earlier work on the Jill Watson project (Goel & Polepeddi 2018) that automatically answered students' questions on discussion forums of online and hybrid classes. One of the main reasons for the success of Jill Watson is that it took a very human-centric approach: it was trained to answer questions that students had actually asked in online discussion forums over a few years. However, Jill Watson answered questions based on course materials such as class syllabi and schedule, by answering questions based on VERA's Users Guide, AskJill generalizes the approach.

## Summary and Conclusions

Explanation of an AI agent requires knowledge of its design and operation. However, acquiring, representing, accessing and using this design knowledge for generating explanations is challenging. In this paper, we noted that almost all practical AI products and services come with a User's Guide that explains both how the product works and how to use the product. This is especially true for AI agents that actually get fielded in real settings and used by real users. Thus, we described the design of a question-answering agent (AskJill) that relies on the Users' Guide to an interactive learning environment (VERA) to explain its domain, functioning and operation. This means that general explanations of the design of an AI agent now can be generated for "free", without requiring any special encoding of knowledge of the agent's design. The general explanations also provide insights into the agent's domain knowledge as well as its mechanisms for achieving its functions.

**Acknowledgements**: AskJill in VERA uses IBM's Watson platform for intent classification. We thank IBM for its support for our work on the AskJill in VERA project.